\documentclass[runningheads]{llncs}

\usepackage{graphicx,paralist,multirow,caption}
\usepackage{comment}
\usepackage{amsmath,amssymb}
\usepackage{color, xcolor, colortbl, tabu}
\usepackage{sidecap}
\usepackage{url}
\usepackage{booktabs}
\DeclareMathOperator{\E}{\mathbb{E}}
\DeclareMathOperator{\R}{\mathbb{R}}
\begin{document}
\pagestyle{headings}
\mainmatter

\def\ACCV20SubNumber{723}  

\title{TinyGAN: Distilling BigGAN for Conditional Image Generation} 
\titlerunning{TinyGAN}
%
\author{Ting-Yun Chang\inst{1}\orcidID{0000-0003-2198-1170} \and
Chi-Jen Lu\inst{1}\orcidID{0000-0003-0835-1190}}
\authorrunning{Chang et al.}
%
\institute{Institute of Information Science, Academia Sinica, Taiwan\\
\email{r06922168@ntu.edu.tw, cjlu@iis.sinica.edu.tw.}\\
\url{https://www.iis.sinica.edu.tw/en/index.html}}

\maketitle

\begin{abstract}
Generative Adversarial Networks (GANs) have become a powerful approach for generative image modeling. However, GANs are notorious for their training instability, especially on large-scale, complex datasets. While the recent work of BigGAN has significantly improved the quality of image generation on ImageNet, it requires a huge model, making it hard to deploy on resource-constrained devices. To reduce the model size, we propose a black-box knowledge distillation framework for compressing GANs, which highlights a stable and efficient training process. Given BigGAN as the teacher network, we manage to train a much smaller student network to mimic its functionality, achieving competitive performance on Inception and FID scores with the generator having $16\times$ fewer parameters.\footnote{The source code and the trained model are publicly available at~\url{https://github.com/terarachang/ACCV_TinyGAN}.}
\end{abstract}

\section{Introduction}
\label{sec:introduction}
Generative Adversarial Networks (GANs)~\cite{goodfellow2014generative} have achieved considerable success in recent years. The framework consists of a generator, which aims to produce a distribution similar to a target one, as well as a discriminator, which aims to distinguish these two distributions. The generator and the discriminator are trained in an alternative way, with the discriminator acting as an increasingly scrupulous critic of the current generator.
Conditional GANs (cGANs)~\cite{mirza2014conditional} are a type of GANs for generating samples based on some given conditional information. Different from unconditional GANs, the discriminator of cGANs is now asked to distinguish the two distributions given the conditional information.

Despite their success, GANs are also known to be hard to train, especially on large-scale, complex datasets such as ImageNet. The recent work of BigGAN~\cite{brock2018large}, a kind of cGANs, demonstrates the benefit of scaling. More precisely, by scaling up both the model size and batch size, some of the training problems can be mitigated, and high-quality images can be generated. 
However, this also leads to high computational cost and memory footprint, even for inference in test time.

\begin{figure}[t!]
    \centering
    \begin{minipage}{1\linewidth}
    \includegraphics[width=1\linewidth]{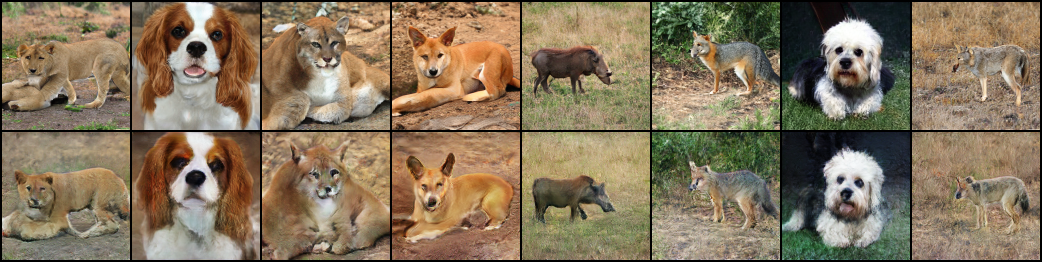}
    \end{minipage}
   
    \begin{minipage}{1\linewidth}
    \includegraphics[width=1\linewidth]{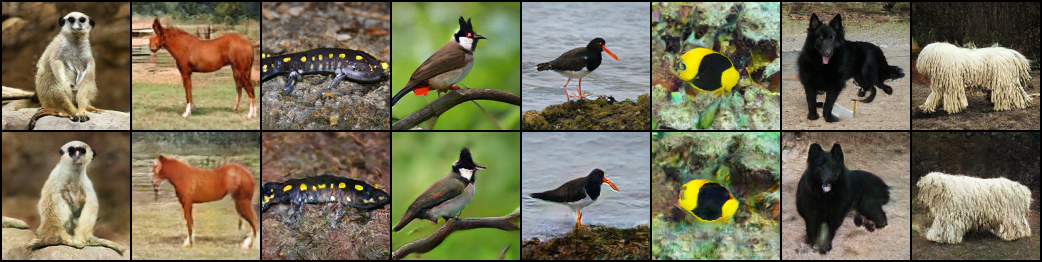}
    \end{minipage}
    
    \caption{A comparison between images generated by BigGAN and the proposed TinyGAN. Pictures in odd rows are produced by BigGAN, while those in even rows are by TinyGAN given the same input.}
    \label{fig:recon_sample}
\end{figure}

One may wonder if it is possible to compress such a large model into a much smaller one. For classification tasks, several techniques have been developed for compressing classifiers, including \textit{knowledge distillation}~\cite{romero2014fitnets}, \textit{network pruning}~\cite{han2015deep}, and \textit{quantization}~\cite{gong2014compressing}.
For compressing GANs, we find that the concept of knowledge distillation (KD) becomes especially appealing. 
Based on a \textit{teacher-student framework}, it aims to impart knowledge encoded in a large, well-trained teacher network to a small student network. For GANs, we find it appropriate to consider the input-output relationship of the teacher generator as the knowledge to be distilled.
Note that the difficulties of training GANs from scratch may be attributed mostly to the lack of supervision from paired training data. Not sure about what the ideal functionality it should have, the generator turns to chase a moving target provided by an evolving discriminator. 
On the other hand, having a well-trained generator such as BigGAN as a teacher, we can use it simply as a black box to generate its input-output pairs as training data, and train a student network in a supervised way. Such a supervised learning is typically much easier, with a much more stable and efficient training process.
In contrast, training classifiers are usually done in a supervised way already, and hence KD on classifiers usually takes a white-box approach, requiring access to the internal of the teacher networks. 

\begin{SCfigure}
  \includegraphics[width=0.55\linewidth]{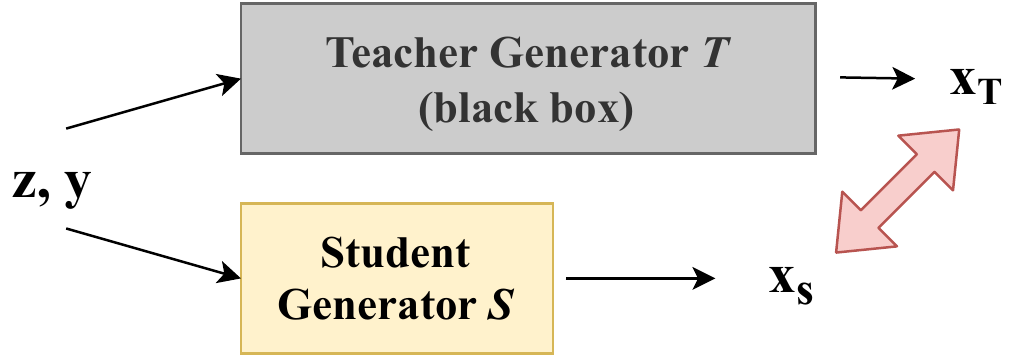}
  
  \caption{Illustration of the problem formulation. $z$ is the noise vector, and $y$ is the class label. Our goal is to mimic the functionality of the teacher generator via black-box knowledge distillation.}
  \label{fig:formula}
\end{SCfigure}

Although KD has been successfully applied to classification tasks~\cite{hinton2015distilling,romero2014fitnets}, it is less studied for image generation.
In our work, we leverage BigGAN trained on ImageNet as our teacher network and design a compact, lightweight student network to mimic the functionality of BigGAN. Given a noise vector and a class label as input, we would like the student network to generate a similar, high-quality image like that produced by BigGAN. In this paper, we focus on \emph{black-box KD}, defined as having access to only the input-output functionality of the teacher network, instead of any internal knowledge such as its intermediate features as needed in works such as \cite{romero2014fitnets,zagoruyko2016paying}. 
We claim that this is a meaningful setting for several aspects. First, it allows one to utilize a model without needing the authority to access its model parameters, by simply collecting its input/output pairs. Next, it allows us to discard the teacher network (both generator and discriminator) in the training phase to save memory after collecting such pairs in the preprocessing step.
Furthermore, it allows us to adopt a different architecture for the student network, which enables us to substantially reduce the model size from that of BigGAN. 
Figure~\ref{fig:formula} is an illustration of our problem formulation, and Figure~\ref{fig:recon_sample} shows some sampled results.

We propose several training objectives for distilling BigGAN, including \emph{pixel-level distillation}, \emph{adversarial distillation}, and \emph{feature-level distillation}. Given the same input, let $x_T$ and $x_S$ be the images generated by the teacher and the student networks respectively. The objective of \emph{pixel-level distillation} is to minimize the distance between $x_T$ and $x_S$, and here we use the pixel-wise L1 distance. We further utilize a small discriminator to help align our generated distribution to BigGAN's, with the \emph{adversarial distillation} having a similar objective as in standard GAN training, but now taking BigGAN's output distribution as the target one. Finally, as pixel-level distance often leads to blurry images, we apply \emph{feature-level distillation} to mitigate this problem. We achieve this without needing additional parameters, by taking the intermediate features in the discriminator and encouraging those derived from $x_S$ to match those from $x_T$.
In addition to the distillation objectives, we also include the standard cGANs loss, to push our generated distribution towards that of ImageNet as well. Our main contributions are summarized as follows.
\begin{itemize}
\item We identify a unique and advantageous property of compressing GANs via knowledge distillation, and initiate the study on the diverse ImageNet.
\item We propose a black-box KD framework tailored to GANs, which requires little permission to the teacher networks and highlights an efficient training process.
\item Our strategy greatly compresses BigGAN, while our model maintains competitive performance.
\end{itemize}
We see our contributions as more conceptual than technical. While the task of compressing classifiers has received much attention, to our knowledge, we are the first to explore black-box KD for compressing GANs. Moreover, we identify a unique property of KD on GANs, which enables us to apply rather simple techniques to achieve a substantial compression ratio, and we believe that it is possible to combine our approach with other compression techniques to further reduce the model size. Let us remark that the emphasis of our work is the realization of a simple and efficient strategy to obtain a generator with both good quality and a compact size. Whereas we do not rule out the possibility of training a small-sized, well-performed, and stable GANs from scratch, it is likely to be challenging except for very skilled and experienced experts. In fact, in our attempt to directly train a smaller GAN from scratch, we have encountered those notorious training problems as expected, while we have never experienced any issues of instability when taking our KD approach. Therefore, our work suggests a possibly more reliable way to obtain a lightweight, high-quality generator: instead of directly training one from scratch, one could first train a large generator and then distill from it a small one.

\section{Related Work}
\paragraph{Generative Adversarial Networks.}
GANs have excelled in a variety of image generation tasks~\cite{pix2pix2017,reed2016generative,ledig2017photo}. Still, they are well known for problems such as training instability and sensitivity to hyperparameter choices, requiring great efforts in model tuning. Several works~\cite{che2016mode,zhao2016energy,arjovsky2017wasserstein,gulrajani2017improved,salimans2018improving} have aimed to tackle such problems. 
Notably, the recent work of~\cite{miyato2018cgans,miyato2018spectral} proposed to constrain the Lipschitz constant of the discriminator function by limiting the spectral norm of its weights, which makes possible 
high quality class-conditional image generation over large-scale, complex distributions.

\paragraph{BigGAN.}
BigGAN~\cite{brock2018large} further scales up GANs by training with considerable model size and batch size on complex datasets.
It basically follows previous SOTA architectures~\cite{miyato2018cgans,miyato2018spectral,zhang2018self}, and proposes two variants, BigGAN and BigGAN-deep, to incorporate the input noises and class labels. Utilizing the truncation trick, i.e., training a model with $z \sim N (0, I)$ but sampling $z$ from a truncated normal (with values falling outside two standard deviations being re-sampled) in test time, BigGAN is able to trade off variety and fidelity. BigGAN demonstrates that GANs benefit dramatically from scaling.

\paragraph{Knowledge Distillation on GANs.}
Perhaps the work most related to ours is~\cite{aguinaldo2019compressing}, which to our knowledge is the first to apply knowledge distillation on GANs. However, their experiments are conducted on MNIST, CIFAR-10, and
CelebA, which are relatively simple with much less image diversity compared to ImageNet. Besides, there are several differences in the settings. First, they do not experiment on conditional generation. Second, they explore teacher-student generators only on the DCGAN architectures, which might be less general and seems easier for the student to mimic a teacher with a similar architecture. Finally, accessing to and updating the teacher discriminator are allowed in their work, while we focus on black-box knowledge distillation, which is more memory efficient during training as we do not need to keep the large teacher network. (Once we synthesized the dataset from BigGAN's generator during the preprocessing phase, we do not need it anymore). To sum up, in this work, we study knowledge distillation on GANs in a more general framework and a harder setting.

\section{Tiny Generative Adversarial Networks}
\label{sec:TinyGAN_obj}
We first describe how our proposed framework, TinyGAN, distills knowledge from BigGAN. Then, we discuss how TinyGAN incorporates real images from the ImageNet dataset, which further improves the performance.

\subsection{BigGAN Distillation}
\label{sec:biggan_distill}
We propose a black-box KD method specifically designed for GANs, which does not need to access the parameters of the teacher network or share a similar network structure.
We use BigGAN as the teacher network and train our student network, TinyGAN, with much fewer parameters to mimic its input-output behavior. We will elaborate on several proposed objectives for knowledge distillation in this subsection.

\begin{figure}[t!]
  \centering
  \includegraphics[width=1.0\linewidth]{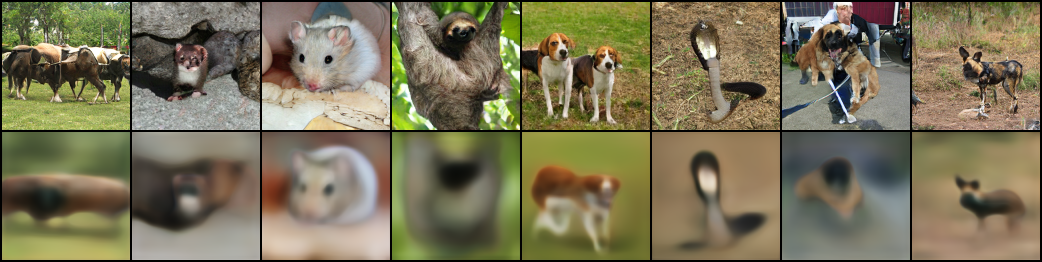}
  \caption{Examples generated by TinyGAN trained with pixel-level distillation loss (Eq.~\ref{eq:loss_pix}) alone, shown in the second row. The first row shows corresponding images produced by BigGAN given the same input.}
  \label{fig:pix_kd_loss_only}
\end{figure}

\paragraph{Pixel-Level Distillation Loss.}
To mimic the functionality of BigGAN, a naive method is to minimize the pixel-level distance between the images generated by BigGAN and TinyGAN given the same 
input. Formally, let
\begin{equation}
\label{eq:loss_pix}
L_\mathrm{KD\_pix} = {\E}_{z \sim p(z), y \sim q(y)} [\| T(z, y) - S(z, y) \|_1],
\end{equation}
where $T$ is the frozen teacher network (BigGAN's generator), $S$ is our student network, $z \in {\R}^{128}$ is a latent variable drawn from the truncated normal distribution $p(z)$, and $y$ is the class label sampled from some categorical distribution $q(y)$. However, we found that using such a pixel-level distance alone is not sufficient for modeling complex datasets such as ImageNet, resulting in blurry images as shown in Figure~\ref{fig:pix_kd_loss_only}. Thus, we propose the following additional objectives to mitigate this problem.

\paragraph{Adversarial Distillation Loss.}
To sharpen the generated images, we incorporate a discriminator to help make the images generated by TinyGAN indistinguishable from those by BigGAN. We adopt an adversarial loss
\begin{equation}\label{eq:loss_kd_adv_G}
L_\mathrm{KD\_S} = -{\E}_{z,y} [D(S(z,y), y)]
\end{equation}
for the generator, and the loss
\begin{equation}
\label{eq:loss_kd_adv_D}
L_\mathrm{KD\_D} = {\E}_{z,y} [\max(0, 1 - D(T(z,y), y)) + \max (0, 1 + D(S(z,y), y))]
\end{equation}
for the discriminator, where $z$ is the noise vector, $y$ is the class label, $T(z, y)$ is the image generated by BigGAN, while $S$ and $D$ are respectively the generator and discriminator of our TinyGAN, which are alternatively trained as in usual GAN training. We trained our small-sized discriminator $D$ from scratch, and have experimentally found that projection discriminator with \textit{hinge} adversarial loss proposed by~\cite{miyato2018cgans} works the best.

\begin{figure}[t!]
  \centering
  \includegraphics[width=1.0\linewidth]{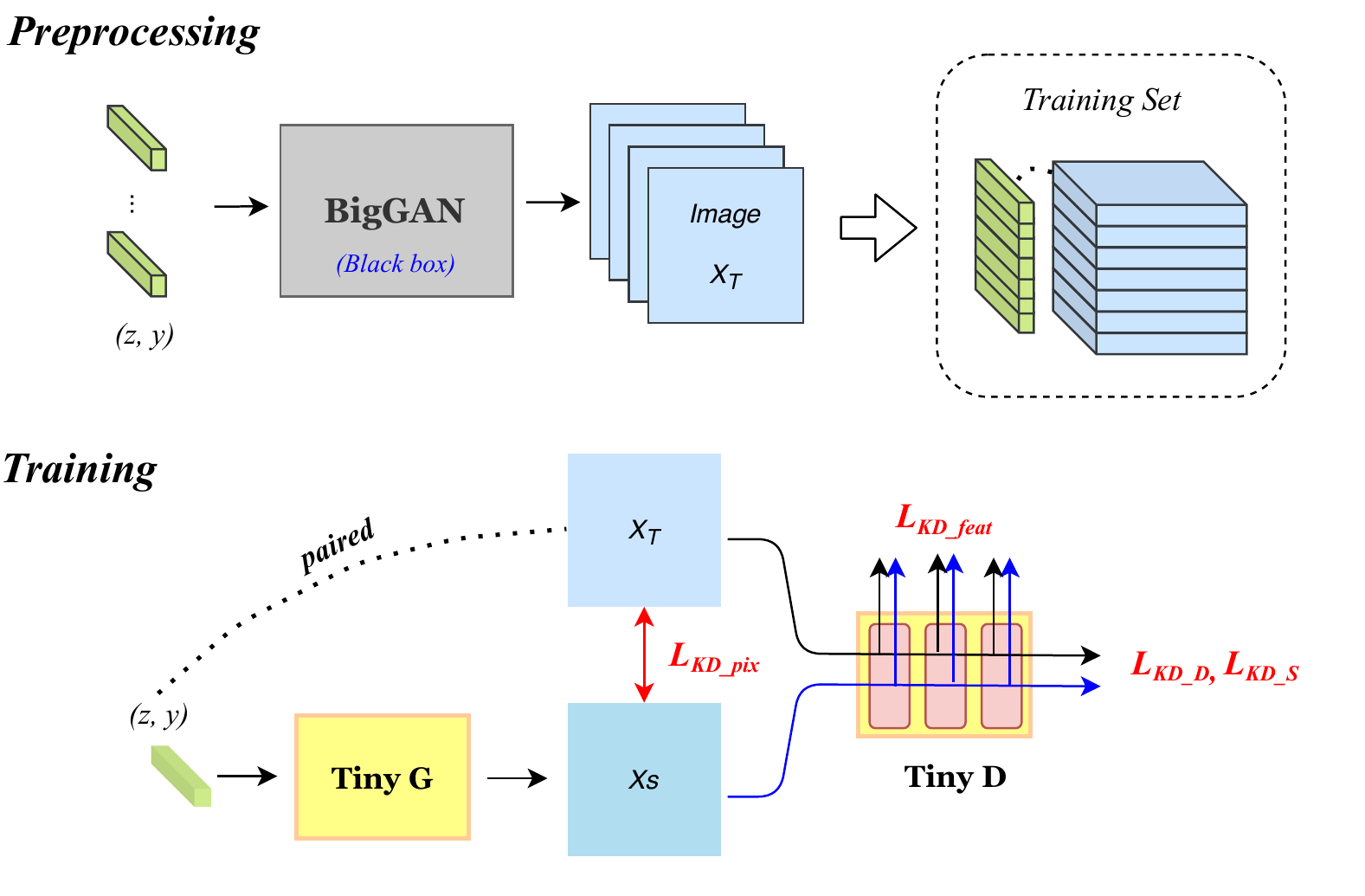}
  \caption{Illustration of the proposed pipeline and the distillation objectives.}
  \label{fig:distill}
\end{figure}

\paragraph{Feature-Level Distillation Loss.}
To further mitigate the problem of generating blurry images using pixel-level distance, we propose a feature-level distillation loss, which does not require any additional parameter. We believe that as the discriminator needs to distinguish the source of images,  it must learn some useful features. Hence, we take the features computed at each convolutional layer in the discriminator, and ask TinyGAN to generate images with similar features as those from BigGAN.
Formally, let
\begin{equation}
\label{eq:loss_feat}
L_\mathrm{KD\_feat} = {\E}_{z,y} [ \Sigma_i \alpha_i \| D_i (T(z, y), y) - D_i (S(z, y), y) \|_1],
\end{equation}
where $D_i$ is the feature vector extracted from the $i$th-layer of our discriminator, and $\alpha_i$ is the corresponding weight. We put more emphasis on higher-level features and assign larger weights to them.

This objective is similar to the feature matching loss proposed by~\cite{wang2018pix2pixHD}, 
which encourages the generator to generate images containing intermediate representations 
similar to those from the real images in order to fool the discriminator.
We have also tried to incorporate different kinds of feature-level loss, such as perceptual loss from VGG network~\cite{simonyan2014very}, but got worse results. Figure~\ref{fig:distill} illustrates all the proposed distillation objectives.

\subsection{Learning from Real Distribution}
We also allow our model to learn from real images in ImageNet dataset, attempting to ameliorate the mode dropping problem of BigGAN we observed in some classes. Specifically, we use the \textit{hinge} version of the adversarial loss~\cite{lim2017geometric}
\begin{equation}
\label{eq:loss_gan}
L_\mathrm{GAN\_D} = {\E}_{x,y} [\max(0, 1 - D(x, y))] + {\E}_{z,y} [\max (0, 1 + D(S(z,y), y))], 
\end{equation}
where $x$ is now the real image sampled from ImageNet. The generator loss $L_\mathrm{GAN\_S}$ is the same as $L_\mathrm{KD\_S}$ in Equation~\eqref{eq:loss_kd_adv_G}.

\subsection{Full Objective} Finally, the objective to optimize our student generator and discriminator, $S$ and $D$, are written respectively as
\begin{equation}
\label{eq:loss_full_G}
L_\mathrm{S} = L_\mathrm{KD\_feat} + {\lambda}_1 L_\mathrm{KD\_pix} + {\lambda}_2 L_\mathrm{KD\_S} + {\lambda}_3 L_\mathrm{GAN\_S}, \mbox{ and }
\end{equation}
\begin{equation}
\label{eq:loss_full_D}
L_\mathrm{D} = L_\mathrm{KD\_D} + {\lambda}_4 L_\mathrm{GAN\_D}.
\end{equation}
Empirically, we gradually decay the weight of the pixel-level distillation loss ${\lambda}_1$ to zero, relying on the discriminator to provide useful guidance. Note that pixel-level distillation loss is still an important term, since it provides stable supervision in the early training phase while discriminator might still be quite naive at that time.

\section{Network Architecture}
Now we describe the architectures of our generator and discriminator in detail.
\subsection{Generator} 
We have tried different generator architectures and experimentally found that ResNet~\cite{he2016identity} based generator with class-conditional BatchNorm~\cite{dumoulin2016learned,de2017modulating} works better. 
To keep a tight computation budget, our student generator does not adopt attention-based~\cite{zhang2018self} or progressive-growing mechanisms~\cite{karras2017progressive}.
To substantially reduce the model size, we mainly rely on using fewer channels and replacing standard convolution by depthwise separable convolution. In addition, we adopt a simpler way to introduce class conditions which also helps the reduction. Overall, our generator has $16\times$ fewer parameters than BigGAN's, while still capable of generating satisfying images of $128 \times 128$ resolution.

\begin{figure}[t!]
\centering
\begin{minipage}[t]{0.48\textwidth}
\centering
\includegraphics[width=5cm]{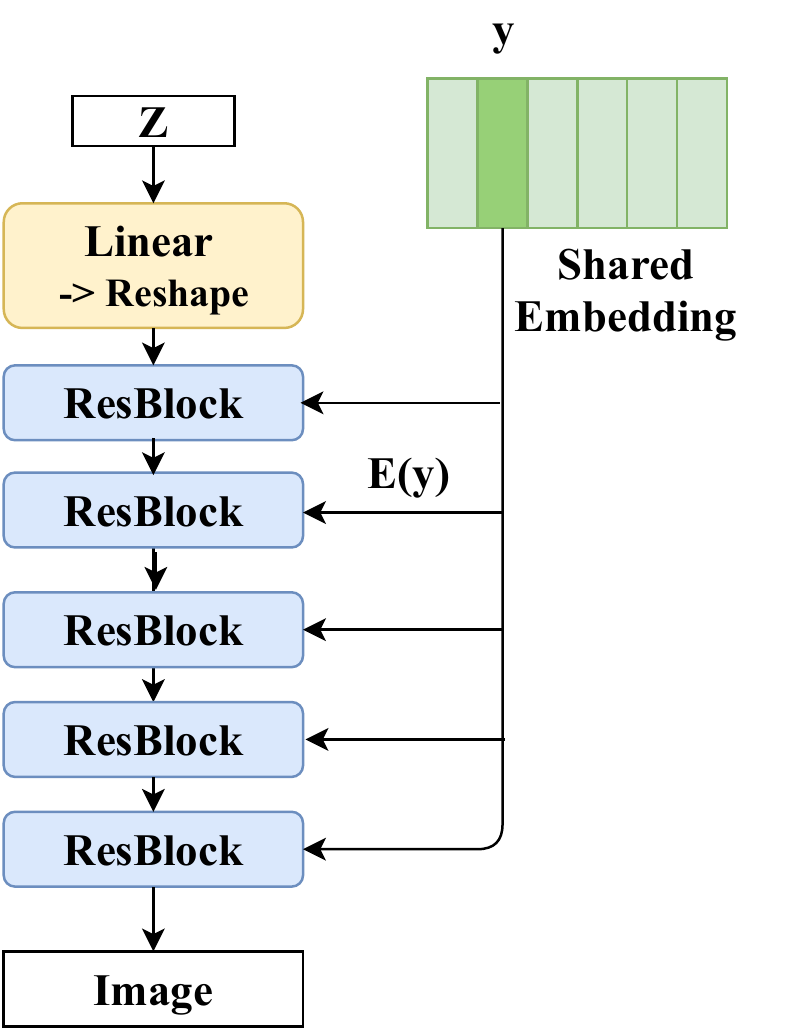}
\caption{Student Generator $S$}
\label{fig:tiny_G_archi}
\end{minipage}
\begin{minipage}[t]{0.48\textwidth}
\centering
\includegraphics[width=6cm]{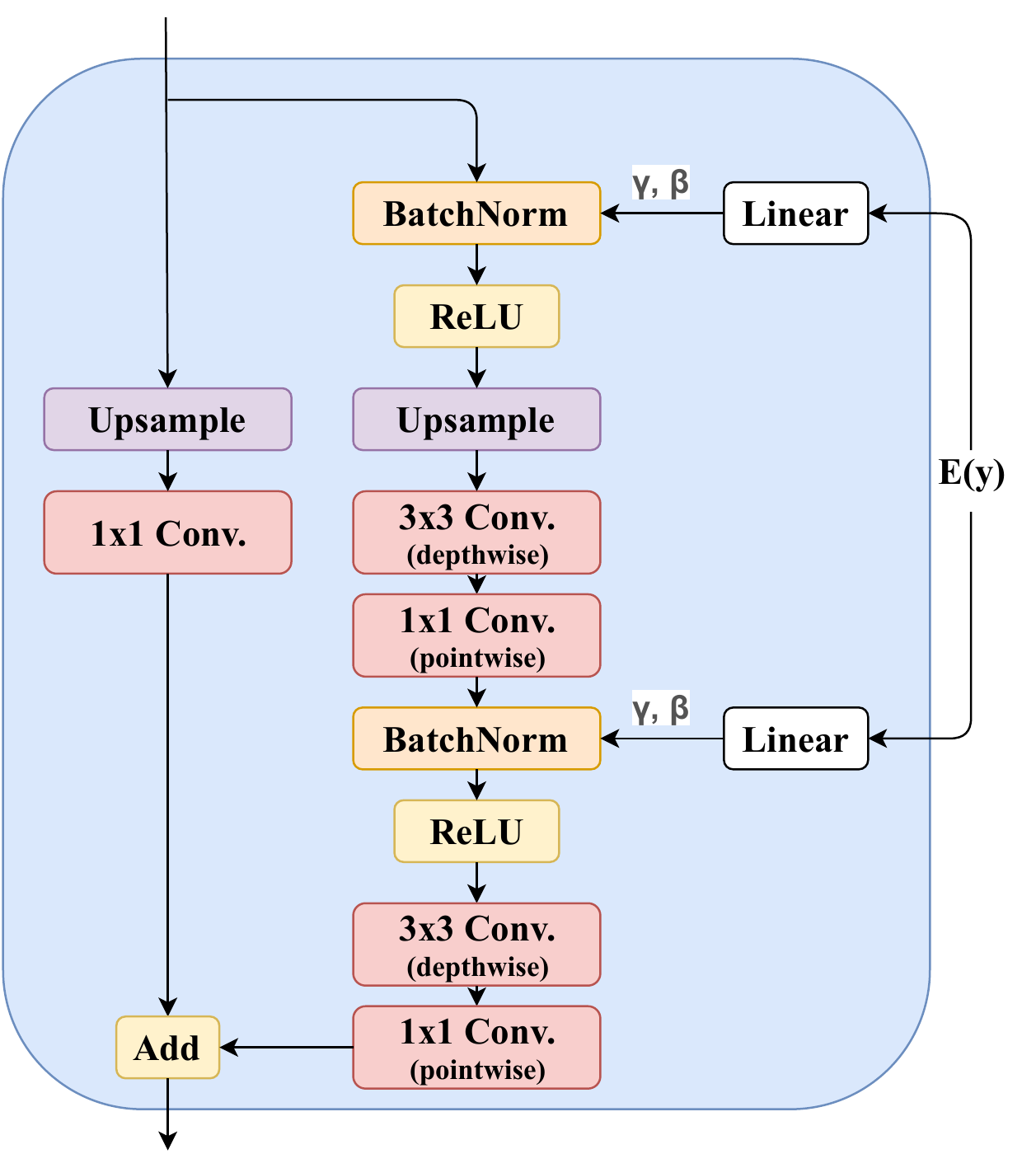}
\caption{A Residual Block in $S$}
\label{fig:G_res}
\end{minipage}
\end{figure}

\paragraph{Shared Class Embedding.} We provide
class information to the generator with class-conditional BatchNorm~\cite{dumoulin2016learned,de2017modulating}.
To reduce computation and memory costs, similar to BigGAN, we use shared class embedding for different layers, which is linearly transformed to produce the BatchNorm affine parameters~\cite{perez2018film}. Different from BigGAN, we design a simpler architecture to incorporate the class label. Specifically, we only input the noise vector $z$ to the first layer, and then for each conditional BatchNorm layer, we linearly transform the class embedding $E(y)$ to the gains and biases. Figure~\ref{fig:tiny_G_archi} is the illustration of our generator architecture. 

\paragraph{Depthwise Separable Convolution.} To further reduce the model size, we replace all the $3 \times 3$  standard convolutional layers in our generator with depthwise separable convolution~\cite{howard2017mobilenets}, which factorizes a standard convolution into a depthwise convolution and a pointwise convolution, by first applying a single filter to each input channel (depthwise), and then utilizing a $1 \times 1$ convolution to combine the outputs (pointwise). Depthwise separable convolution uses $\frac{1}{O} + \frac{1}{k \times k}$ fewer parameters than the standard one, where $O$ is the number of output channels and $k$ is the kernel size. We denote TinyGAN using standard conv. layers as TinyGAN-std, and the variant with depth-wise conv. layer as TinyGAN-dw.\footnote{Note that all the figures in this paper are generated by the TinyGAN-dw variant.}


\subsection{Discriminator}
With the supervision from BigGAN, the difficulties of training is greatly reduced and we found that a simple discriminator architecture already works well. Following ~\cite{miyato2018spectral,miyato2018cgans}, we use spectral normalized discriminator and introduce the class condition via projection. But instead of utilizing complicated residual blocks, we found that simply stacking multiple convolutional layers with stride as DCGAN~\cite{radford2015unsupervised} works well enough, which greatly reduces the number of parameters.
In fact, our discriminator is $10\times$ smaller than that of BigGAN's.

\section{Experiments}
\subsection{Datasets}
\paragraph{ImageNet.}
The ImageNet ILSVRC 2012 dataset~\cite{russakovsky2015imagenet} consists of 1,000 image classes, each having approximately 1,300 images. We compressed each image to 128x128 pixels, using the source code released by \cite{miyato2018cgans}.
\paragraph{Images Generated by BigGAN.}
We view BigGAN, our teacher network, as a black-box model and collect its input-output pairs to train our student network. For each class, we randomly sample 3,000\footnote{We also tried $1000$ instances per class, which already achieves good results; however, no significant improvement was observed when we increased to $4000$.} noise vectors from the truncated normal distribution and collect the corresponding output generated by BigGAN using the official API.\footnote{\url{https://tfhub.dev/deepmind/biggan-deep-128/1}}

As we found TinyGAN unable to model some complicated objects well enough, 
we only report in Table~\ref{table:exp} the IS/FID/intra-FID scores measured on \emph{all} animal classes (398 classes in total). It shows that our approach can work well for a large set of homogeneous classes, and we focus on animals as they may have more downstream applications than other classes. Further discussion about experiments on all 1000 classes can be found in the supplementary material.

\subsection{Evaluation Metrics}
\paragraph{Inception score (IS).}
IS~\cite{salimans2016improved} measures the KL-divergence
between the conditional class distribution $p(y|x)$ and the marginal class distribution $p(y)$. Formally,
\begin{equation}
    \mathrm{IS} = \exp{(\E_{x}[\mathrm{KL}(p(y|x) \| p(y))])}.
\end{equation}
A higher Inception score suggests a better performance. Despite the limitations of IS~\cite{lucic2018gans}, we still adopt it as it is widely used in prior works.

\paragraph{Fréchet Inception Distance (FID).}
FID score~\cite{heusel2017gans} computes the 2-Wasserstein distance between the two distributions $r$ and $g$, and is given by 
\begin{equation}
\mathrm{FID}(r, g) = \| {\mu}_{r} - {\mu}_{g} \|_2 + \mathrm{Tr}({\Sigma}_{r} + {\Sigma}_{g} - 2({\Sigma}_{r} {\Sigma}_{g})^{1/2}).
\end{equation}
Here, ${\mu}_{r}$ and $\mu_g$ are the means of the final feature vectors extracted from the inception model~\cite{szegedy2015going} with input from the real and generated samples respectively, while $\Sigma_r$ and ${\Sigma}_{g}$ are the corresponding covariance matrices, and Tr is the trace. We also compute the intra-FID score~\cite{miyato2018cgans}, which measures the average FID score within each class.

Unlike Inception score, FID is able to detect intra-class mode dropping. It is considered a more consistent estimator~\cite{lucic2018gans}, as a model that generates only a single image per class can score a perfect IS but not a good FID. Here we follow prior works and use \texttt{TensorFlow toolkit} to calculate IS and FID scores.

\subsection{Baseline Models}
\paragraph{SNGAN-Projection.} 
Spectral Normalization GAN (SNGAN) \cite{miyato2018spectral} proposes spectral normalization to stabilize the training of the discriminator. \cite{miyato2018cgans} further proposes a projection-based discriminator, which incorporates the class labels via inner product instead of concatenation.
The combined model, denoted as SNGAN-Projection, has shown significant improvements on ImageNet Dataset, and we consider it as a strong baseline model. Statistics in Table~\ref{table:exp} are reported using the source code and the pretrained generator released by the authors.\footnote{\url{https://github.com/pfnet-research/sngan_projection}} 

\paragraph{SAGAN.}
Self-Attention GAN (SAGAN)~\cite{zhang2018self} is built atop SNGAN-projection, and introduces a self-attention mechanism~\cite{parikh2016decomposable,vaswani2017attention} into convolutional GANs, in order to model long-range dependencies across image regions. As the authors do not provide a pretrained model and we are unable to train it from scratch due to limits of computation, its scores are left blank in Table~\ref{table:exp}, and we only compare its model size and computation cost to our TinyGAN. For reference, the IS/FID/intra-FID scores reported in the original paper, evaluated on all 1000 classes, are $\bf 52.52/ 18.7/ 83.7$ respectively. 

\paragraph{TinyGAN trained from scratch.}
To justify the effectiveness of knowledge distillation on GANs, we also experimented on training TinyGAN from scratch, without the guidance of a teacher network. That is, we use an identical architecture of TinyGAN, but trained it using only the adversarial loss $L_\mathrm{GAN}$ (Eq.~\ref{eq:loss_gan}).

\subsection{Training}
The proposed TinyGAN are trained using Adam~\cite{kingma2014adam} with $\beta_1 = 0.0$ and $\beta_2 = 0.9$. The learning rates of generator and discriminator are both set to $0.0002$ with linear decay. We perform one generator update after $10$ discriminator updates. With the stable guidance from the teacher network, no special tricks for training GANs are needed. While BigGAN notes that using a large batch size boosts the performance, in our TinyGAN, we found that a smaller batch size (32 or 16) works as well. Training takes about 3 days on a single NVIDIA 2080Ti GPU.

\begin{table}[t!]
\setlength{\tabcolsep}{4pt}
\small
  \centering
  \begin{tabu}{lcrrrccc}
    \toprule
    \bf Model & \bf Ch. & \bf \#Par. & \bf G Par. & \bf FLOPs & \bf{IS} $\uparrow$ & \bf{FID} $\downarrow$ & \bf{intra-FID} $\downarrow$ \\
    \hline\hline 
    SNGAN-proj & 64 & 72.0 M & 42.0 M & 9.10 B & $ 31.4 \pm 0.7 $ & 29.0 & 84.1 \\
    SAGAN & 64 & 81.5 M & 42.0 M & 9.18 B & - & - & - \\
    BigGAN-deep & 128 & 85.0 M & 50.4 M & 8.32 B & $\bf{146.1 \pm 1.7}$& \bf{19.8} & \bf{55.6} \\
    \hline
    TinyGAN-std & 32 & 12.6 M & 9.3 M & 2.29 B & $ {94.0 \pm 1.2}$ & 21.6 & 70.6 \\ 
    \rowfont{\color{gray}} TinyGAN-std & 16 & 6.2 M & 2.9 M & 0.58 B & $ {68.25 \pm 1.0}$ & 27.4 & 88.1 \\ 
    TinyGAN-dw & 32 & 6.4 M & 3.1 M & 0.44 B & $79.19 \pm 1.6$ & 24.2 & 79.1\\
    \bottomrule
  \end{tabu}
  \vspace{+1mm}
  \caption{ Inception Score (IS, higher is better) and Fréchet Inception Distance (FID, lower is better). Ch. is the channel multiplier representing the number of units in each layer. \#Par. is total number of parameters. We highlight the generator's parameters G Par. since the discriminator is not required for inference. M denotes million and B is billion.} 
  \label{table:exp}
\end{table}

 \begin{figure}[t!]
  \centering
  \begin{minipage}[t]{0.48\textwidth}
  \includegraphics[width=1.0\linewidth]{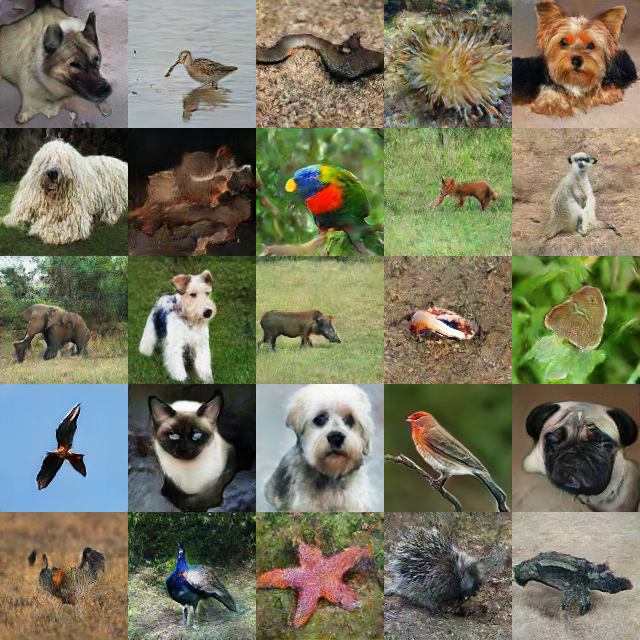}
 \end{minipage}
 \begin{minipage}[t]{0.48\textwidth}
  \includegraphics[width=1.0\linewidth]{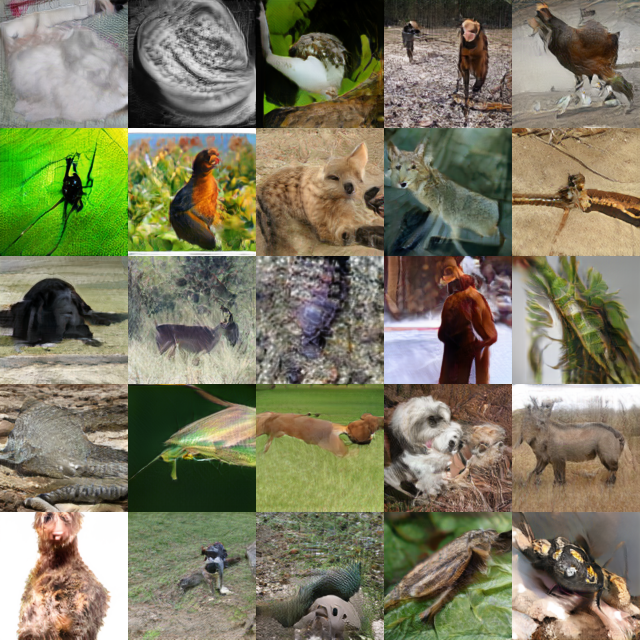}
  \end{minipage}
  \caption{A comparison between randomly sampled images generated by TinyGAN-dw (left) and SNGAN-projection (right).}
\label{fig:compare_sngan}
\end{figure}

\subsection{Results}
We evaluate TinyGAN on all the 398 animal classes in the ImageNet dataset, and the results are shown in Table~\ref{table:exp}. We compare the computation cost of TinyGAN with the teacher network (BigGAN-deep) and two strong baseline models discussed before. 
 
Note that our proposed model uses 
much fewer parameters and floating-point operations than all the other frameworks. We also study the trade-off between model size and performance of different variants of TinyGAN in the last three rows in Table~\ref{table:exp}. Experiments show that TinyGAN with standard conv. layers (TinyGAN-std) achieves the best performance but uses more parameters. To reduce the model size, we can either reduce the channel multiplier or adopting depth-wise separable conv. layers (TinyGAN-dw). The result shows that aggressively reducing channels leads to a noticeable drop in performance. On the other hand, it is much less significant in TinyGAN-dw, making it a suitable choice under a tight computation budget. Specifically, our generator of TinyGAN-std/TinyGAN-dw has $\sim 18\%/ 6\%$ parameters and $\sim 28\%/ 5\%$ FLOPs when compared with the teacher network, and we also have
similar reductions from the other two baseline models.

Although there is a performance gap between our TinyGAN-dw and the teacher network, we claim that it is tolerable considering its compact model size, and its better performance over SNGAN-projection in all the metrics. We further compare the image quality of TinyGAN-dw and SNGAN-projection in Figure~\ref{fig:compare_sngan}, where all images are randomly sampled within animal classes. We found that while SNGAN-projection is able to produce sharper images with clear details, perhaps due to its larger model complexity, our TinyGAN focuses on the intended class itself and generates more realistic images with less distortion.\footnote{More randomly sampled images for comparisons between TinyGAN and SNGAN-projection can be found in the supplementary material.}

 \begin{table}[t!]
  \setlength{\tabcolsep}{12pt}
  \centering
  \begin{tabular}{lccccccc}
    \toprule
    \bf Model & \bf{FID} $\downarrow$ & \bf{intra-FID} $\downarrow$ \\
    \hline\hline 
    TinyGAN-dw & 24.2 & 79.1 \\
    $- L_\mathrm{KD\_feat}$ & 54.4 & 149.2 \\
    $- L_\mathrm{GAN}$ & 28.8 & 89.9 \\
    $- L_\mathrm{KD\_S}, L_\mathrm{KD\_D}$ & 60.5 & 157.0 \\
    \hline
    $ \enspace L_\mathrm{KD\_pix}$ & 107.9 & 216.0 \\
    \bottomrule
  \end{tabular}
  \caption{Ablation Study} 
  \label{table:ablation}
\end{table}

\begin{figure}[t!]
\begin{minipage}[t]{0.48\textwidth}
  \centering
  \includegraphics[width=1.0\linewidth]{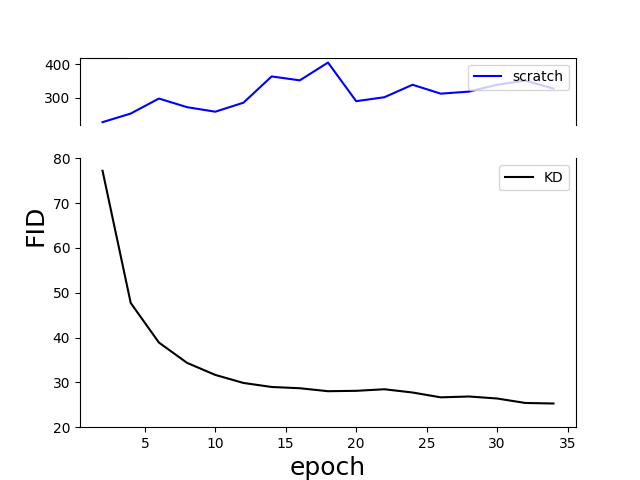}
  \caption{FID scores during the training phase of TinyGAN-dw (trained from scratch v.s. trained with KD losses).}
  \label{fig:learning_curve}
\end{minipage}
\hfill
\begin{minipage}[t]{0.48\textwidth}
  \centering
  \includegraphics[width=1.0\linewidth]{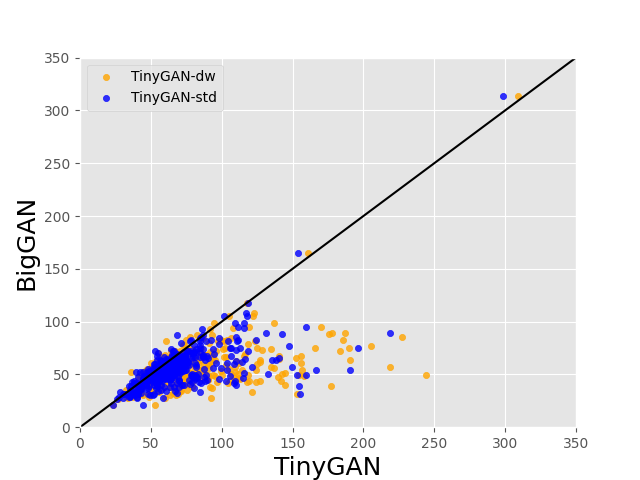}
  \caption{Comparing intra-FID scores of TinyGAN and BigGAN. Each dot corresponds to a class.}
  \label{fig:correlation}
\end{minipage}
\end{figure}

Finally, let us stress that the main point of our work is not to claim how small our network is, but to propose an easy way to train such one. Figure~\ref{fig:learning_curve} shows the learning curve of TinyGAN-dw, demonstrating a smooth, stable and efficient training process it has. In fact, with our knowledge distillation losses, we have never experienced any training collapse, and most of our effort has been spent on finding the right balance between model size and image quality. On the other hand, we have also experimented on training TinyGAN from scratch, and the blue line in Figure~\ref{fig:learning_curve} shows a typical training failure we often encountered. Although we do not rule out the possibility of training a small-size GAN from scratch, based on the network architecture of either TinyGAN, BigGAN, or other baselines, a successful training is likely to be hard without considerable efforts for overcoming those well-known training problems. 

\subsection{Analysis}
 \begin{figure}[t!]
  \centering
  \begin{minipage}[t]{0.48\textwidth}
  \includegraphics[width=1.0\linewidth]{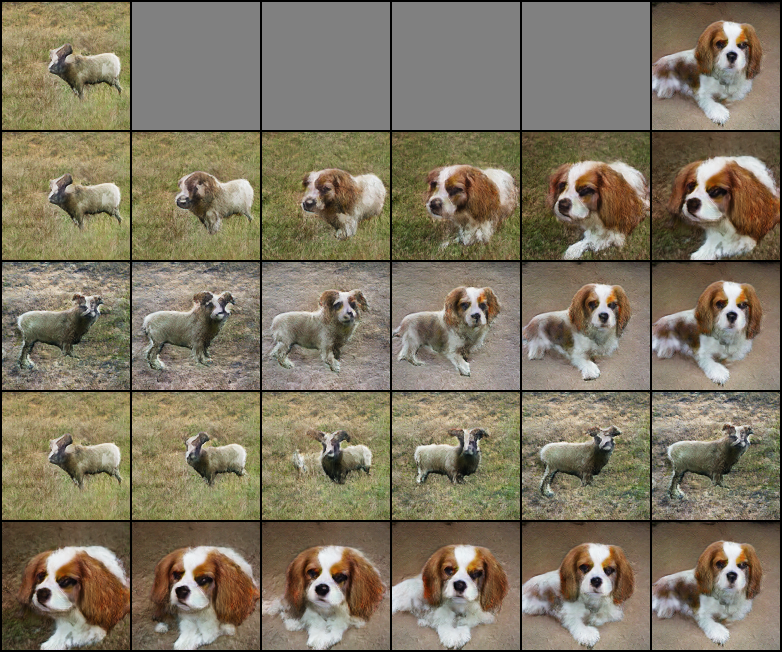}
 \end{minipage}
 \begin{minipage}[t]{0.48\textwidth}
  \includegraphics[width=1.0\linewidth]{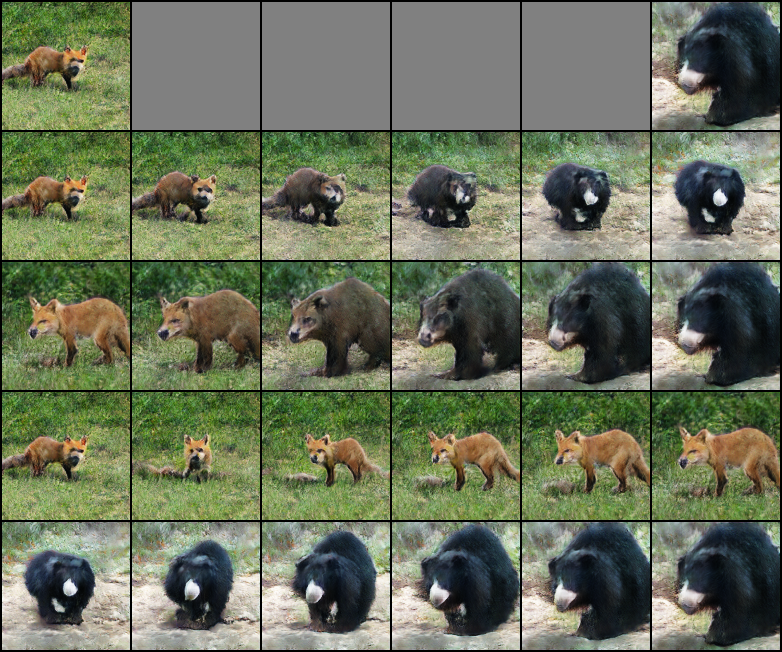}
  \end{minipage}
  \caption{Interpolations between $z, y$ pairs. The second and third rows interpolate between $y$ with $z$ fixed, while the last two rows interpolate between $z$ with $y$ fixed. Observe that semantics are maintained
between two endpoints.}
\label{fig:interpolation}
\end{figure}

\paragraph{Ablation Study.}
We conduct ablation study to validate those objectives proposed in Section~\ref{sec:biggan_distill}. Table~\ref{table:ablation} shows the results of omitting 
$L_\mathrm{KD\_feat}$ (Eq.~\ref{eq:loss_feat}), $L_\mathrm{GAN}$ (Eq.~\ref{eq:loss_gan}), and $L_\mathrm{KD\_S}, L_\mathrm{KD\_D}$ (Eq.~\ref{eq:loss_kd_adv_G},\ref{eq:loss_kd_adv_D}) respectively, as well as that of using 
$L_\mathrm{KD\_pix}$ (Eq.~\ref{eq:loss_pix}) alone without the discriminator.

The fifth row in Table~\ref{table:ablation} shows that adding a discriminator, which only costs a few parameters, is very crucial for the performance. 
Because the discriminator is trained to discern real from fake images, it guides the generator to produce sharper and more realistic images.
Similarly, feature-level distillation loss $L_\mathrm{KD\_feat}$ significantly improves the performance as the generator learns to match the informative features extracted from the discriminator. In addition, omitting adversarial distillation loss $L_\mathrm{KD\_S}, L_\mathrm{KD\_D}$ and keeping the others is equivalent to training a standard cGANs (with the real distribution from ImageNet) while incorporating supervision from the teacher via pixel-wise and feature-wise losses.
The notable drop in performance in the fourth row indicates the importance of leveraging the discriminator to push the student's output distribution to the teacher's.

In addition to the distillation objectives which provide stable supervision from the teacher, including the standard adversarial loss $L_\mathrm{GAN}$ further improves the image quality. The ablation study demonstrates that all the proposed objectives in Section~\ref{sec:TinyGAN_obj} are useful for training our TinyGAN.

\paragraph{Interpolation.}
To understand the generalization ability of our TinyGAN-dw, we perform linear interpolations between random noise vectors $z_1, z_2$ and class labels $y_1$, $y_2$. 

We first interpolate between the class embedding $E(y_1)$ and $E(y_2)$ with the noise vectors fixed. In Figure~\ref{fig:interpolation}, the second and third rows demonstrate that TinyGAN can successfully perform category morphing. We then interpolate between the noise vectors $z_1$ and $z_2$ with fixed class labels. The last two rows in Figure~\ref{fig:interpolation} show that TinyGAN can also smoothly manipulate some coarse features such as poses and sizes of the animals.

\begin{figure}[t!]
    \centering
    \begin{minipage}{1\linewidth}
    \includegraphics[width=1\linewidth]{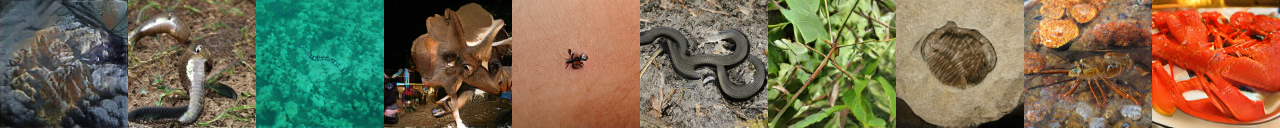}
    \end{minipage}

    \begin{minipage}{1\linewidth}
    \includegraphics[width=1\linewidth]{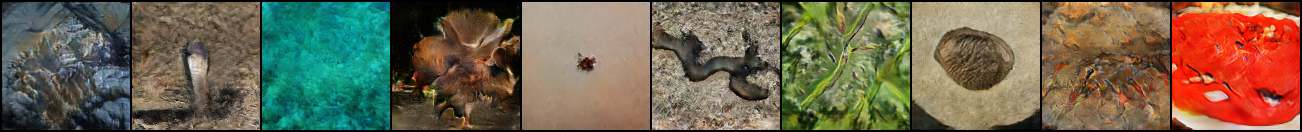}
    \end{minipage}
    \caption{Samples of the 10 worst classes. The first row is generated by BigGAN and the second row is by TinyGAN-dw.}
    \label{fig:bad_classes}
\end{figure}

\paragraph{Quality Analysis.}
Finally, to better understand the weakness of our TinyGAN, we investigate on classes with high intra-FID scores. We first show the positive correlation (Pearson's correlation coefficient $= 0.54, 0.64$) between teacher and student networks (-dw, -std) in Figure~\ref{fig:correlation}, which reveals that TinyGAN's failure in a few classes can be attributed to the teacher network. We then focus on the 10 worst classes with the highest FID scores, which are chambered nautilus, Indian cobra, sea snake, triceratops, tick, ringneck snake, walking stick, trilobite, crayfish, and American lobster. As the samples in Figure~\ref{fig:bad_classes} show, most of them have complicated or delicate appearances and bear little resemblance to most of the others, making them hard to model with others by a small network. 

\section{Conclusion}
Training GANs from scratch has well-known problems, especially for complex datasets such as ImageNet, and the recent work of BigGAN shows that scaling up GANs can mitigate some of the problems and produce high-quality images. However, it requires huge computational resources not only for training but also for testing, which may prevent its use in resource-limited devices. We propose a novel black-box knowledge distillation method for GANs, which allows us to learn a much smaller generator with competitive performance in an efficient and stable way when given a well-trained large generator such as BigGAN.
\bibliographystyle{splncs}
\bibliography{egbib}

\begin{thebibliography}{10}

\bibitem{goodfellow2014generative}
Goodfellow, I., Pouget-Abadie, J., Mirza, M., Xu, B., Warde-Farley, D., Ozair,
  S., Courville, A., Bengio, Y.:
\newblock Generative adversarial nets.
\newblock In: Advances in neural information processing systems. (2014)
  2672--2680

\bibitem{mirza2014conditional}
Mirza, M., Osindero, S.:
\newblock Conditional generative adversarial nets.
\newblock arXiv preprint arXiv:1411.1784 (2014)

\bibitem{brock2018large}
Brock, A., Donahue, J., Simonyan, K.:
\newblock Large scale gan training for high fidelity natural image synthesis.
\newblock ICLR (2019)

\bibitem{romero2014fitnets}
Romero, A., Ballas, N., Kahou, S.E., Chassang, A., Gatta, C., Bengio, Y.:
\newblock Fitnets: Hints for thin deep nets.
\newblock arXiv preprint arXiv:1412.6550 (2014)

\bibitem{han2015deep}
Han, S., Mao, H., Dally, W.J.:
\newblock Deep compression: Compressing deep neural networks with pruning,
  trained quantization and huffman coding.
\newblock ICLR (2016)

\bibitem{gong2014compressing}
Gong, Y., Liu, L., Yang, M., Bourdev, L.:
\newblock Compressing deep convolutional networks using vector quantization.
\newblock ICLR (2015)

\bibitem{hinton2015distilling}
Hinton, G., Vinyals, O., Dean, J.:
\newblock Distilling the knowledge in a neural network.
\newblock arXiv preprint arXiv:1503.02531 (2015)

\bibitem{zagoruyko2016paying}
Zagoruyko, S., Komodakis, N.:
\newblock Paying more attention to attention: Improving the performance of
  convolutional neural networks via attention transfer.
\newblock ICLR (2017)

\bibitem{pix2pix2017}
Isola, P., Zhu, J.Y., Zhou, T., Efros, A.A.:
\newblock Image-to-image translation with conditional adversarial networks.
\newblock CVPR (2017)

\bibitem{reed2016generative}
Reed, S., Akata, Z., Yan, X., Logeswaran, L., Schiele, B., Lee, H.:
\newblock Generative adversarial text to image synthesis.
\newblock ICML (2016)

\bibitem{ledig2017photo}
Ledig, C., Theis, L., Husz{\'a}r, F., Caballero, J., Cunningham, A., Acosta,
  A., Aitken, A., Tejani, A., Totz, J., Wang, Z.,  et~al.:
\newblock Photo-realistic single image super-resolution using a generative
  adversarial network.
\newblock In: Proceedings of the IEEE conference on computer vision and pattern
  recognition. (2017)  4681--4690

\bibitem{che2016mode}
Che, T., Li, Y., Jacob, A.P., Bengio, Y., Li, W.:
\newblock Mode regularized generative adversarial networks.
\newblock ICLR (2017)

\bibitem{zhao2016energy}
Zhao, J., Mathieu, M., LeCun, Y.:
\newblock Energy-based generative adversarial network.
\newblock ICLR (2017)

\bibitem{arjovsky2017wasserstein}
Arjovsky, M., Chintala, S., Bottou, L.:
\newblock Wasserstein gan.
\newblock ICML (2017)

\bibitem{gulrajani2017improved}
Gulrajani, I., Ahmed, F., Arjovsky, M., Dumoulin, V., Courville, A.C.:
\newblock Improved training of wasserstein gans.
\newblock In: Advances in neural information processing systems. (2017)
  5767--5777

\bibitem{salimans2018improving}
Salimans, T., Zhang, H., Radford, A., Metaxas, D.:
\newblock Improving gans using optimal transport.
\newblock ICLR (2018)

\bibitem{miyato2018cgans}
Miyato, T., Koyama, M.:
\newblock cgans with projection discriminator.
\newblock ICLR (2018)

\bibitem{miyato2018spectral}
Miyato, T., Kataoka, T., Koyama, M., Yoshida, Y.:
\newblock Spectral normalization for generative adversarial networks.
\newblock ICLR (2018)

\bibitem{zhang2018self}
Zhang, H., Goodfellow, I., Metaxas, D., Odena, A.:
\newblock Self-attention generative adversarial networks.
\newblock arXiv preprint arXiv:1805.08318 (2018)

\bibitem{aguinaldo2019compressing}
Aguinaldo, A., Chiang, P.Y., Gain, A., Patil, A., Pearson, K., Feizi, S.:
\newblock Compressing gans using knowledge distillation.
\newblock arXiv preprint arXiv:1902.00159 (2019)

\bibitem{wang2018pix2pixHD}
Wang, T.C., Liu, M.Y., Zhu, J.Y., Tao, A., Kautz, J., Catanzaro, B.:
\newblock High-resolution image synthesis and semantic manipulation with
  conditional gans.
\newblock In: Proceedings of the IEEE Conference on Computer Vision and Pattern
  Recognition. (2018)

\bibitem{simonyan2014very}
Simonyan, K., Zisserman, A.:
\newblock Very deep convolutional networks for large-scale image recognition.
\newblock arXiv preprint arXiv:1409.1556 (2014)

\bibitem{lim2017geometric}
Lim, J.H., Ye, J.C.:
\newblock Geometric gan.
\newblock arXiv preprint arXiv:1705.02894 (2017)

\bibitem{he2016identity}
He, K., Zhang, X., Ren, S., Sun, J.:
\newblock Identity mappings in deep residual networks.
\newblock In: European conference on computer vision, Springer (2016)  630--645

\bibitem{dumoulin2016learned}
Dumoulin, V., Shlens, J., Kudlur, M.:
\newblock A learned representation for artistic style.
\newblock ICLR (2017)

\bibitem{de2017modulating}
De~Vries, H., Strub, F., Mary, J., Larochelle, H., Pietquin, O., Courville,
  A.C.:
\newblock Modulating early visual processing by language.
\newblock In: Advances in Neural Information Processing Systems. (2017)
  6594--6604

\bibitem{karras2017progressive}
Karras, T., Aila, T., Laine, S., Lehtinen, J.:
\newblock Progressive growing of gans for improved quality, stability, and
  variation.
\newblock ICLR (2018)

\bibitem{perez2018film}
Perez, E., Strub, F., De~Vries, H., Dumoulin, V., Courville, A.:
\newblock Film: Visual reasoning with a general conditioning layer.
\newblock In: Thirty-Second AAAI Conference on Artificial Intelligence. (2018)

\bibitem{howard2017mobilenets}
Howard, A.G., Zhu, M., Chen, B., Kalenichenko, D., Wang, W., Weyand, T.,
  Andreetto, M., Adam, H.:
\newblock Mobilenets: Efficient convolutional neural networks for mobile vision
  applications.
\newblock arXiv preprint arXiv:1704.04861 (2017)

\bibitem{radford2015unsupervised}
Radford, A., Metz, L., Chintala, S.:
\newblock Unsupervised representation learning with deep convolutional
  generative adversarial networks.
\newblock ICLR (2016)

\bibitem{russakovsky2015imagenet}
Russakovsky, O., Deng, J., Su, H., Krause, J., Satheesh, S., Ma, S., Huang, Z.,
  Karpathy, A., Khosla, A., Bernstein, M.,  et~al.:
\newblock Imagenet large scale visual recognition challenge.
\newblock International journal of computer vision \textbf{115} (2015)
  211--252

\bibitem{salimans2016improved}
Salimans, T., Goodfellow, I., Zaremba, W., Cheung, V., Radford, A., Chen, X.:
\newblock Improved techniques for training gans.
\newblock In: Advances in neural information processing systems. (2016)
  2234--2242

\bibitem{lucic2018gans}
Lucic, M., Kurach, K., Michalski, M., Gelly, S., Bousquet, O.:
\newblock Are gans created equal? a large-scale study.
\newblock In: Advances in neural information processing systems. (2018)
  700--709

\bibitem{heusel2017gans}
Heusel, M., Ramsauer, H., Unterthiner, T., Nessler, B., Hochreiter, S.:
\newblock Gans trained by a two time-scale update rule converge to a local nash
  equilibrium.
\newblock In: Advances in Neural Information Processing Systems. (2017)
  6626--6637

\bibitem{szegedy2015going}
Szegedy, C., Liu, W., Jia, Y., Sermanet, P., Reed, S., Anguelov, D., Erhan, D.,
  Vanhoucke, V., Rabinovich, A.:
\newblock Going deeper with convolutions.
\newblock In: Proceedings of the IEEE conference on computer vision and pattern
  recognition. (2015)  1--9

\bibitem{parikh2016decomposable}
Parikh, A.P., T{\"a}ckstr{\"o}m, O., Das, D., Uszkoreit, J.:
\newblock A decomposable attention model for natural language inference.
\newblock EMNLP (2016)

\bibitem{vaswani2017attention}
Vaswani, A., Shazeer, N., Parmar, N., Uszkoreit, J., Jones, L., Gomez, A.N.,
  Kaiser, {\L}., Polosukhin, I.:
\newblock Attention is all you need.
\newblock In: Advances in neural information processing systems. (2017)
  5998--6008

\bibitem{kingma2014adam}
Kingma, D.P., Ba, J.:
\newblock Adam: A method for stochastic optimization.
\newblock ICLR (2015)

\end{thebibliography}

\clearpage
\appendix

\section{More Samples Generated by TinyGAN}
Here we show more samples and interpolation results generated by the proposed TinyGAN, as well as some samples by SNGAN-projection for comparison.

\begin{figure}[b!]
    \centering
    \begin{minipage}{1\linewidth}
    \includegraphics[width=1\linewidth]{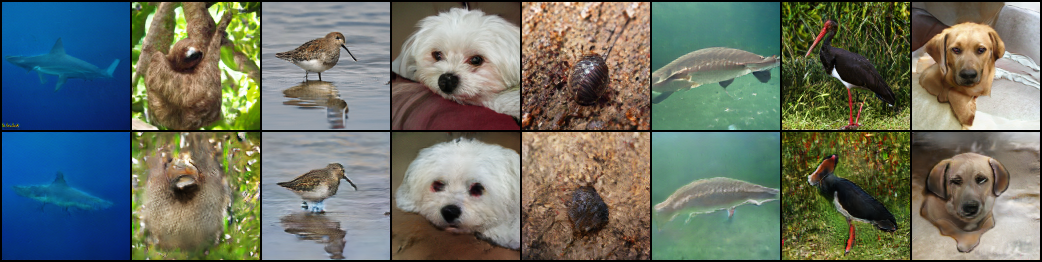}
    \end{minipage}
    
    \begin{minipage}{1\linewidth}
    \includegraphics[width=1\linewidth]{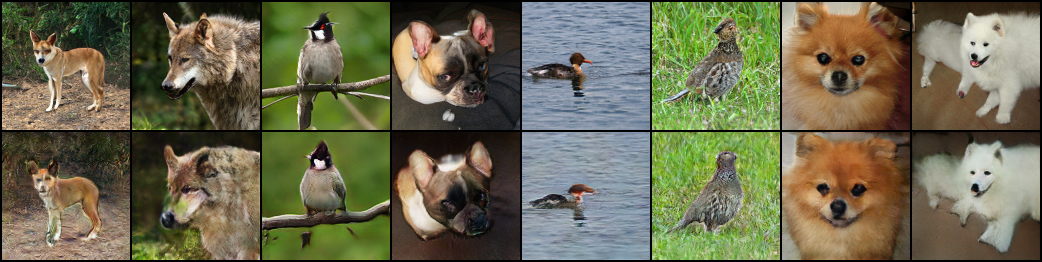}
    \end{minipage}
   
    \begin{minipage}{1\linewidth}
    \includegraphics[width=1\linewidth]{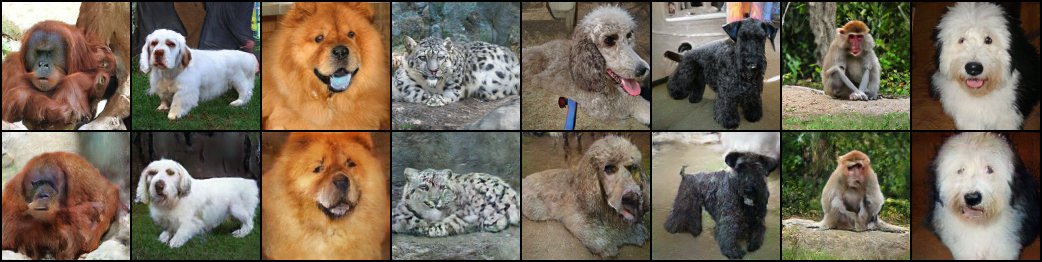}
    \end{minipage}
    
    \begin{minipage}{1\linewidth}
    \includegraphics[width=1\linewidth]{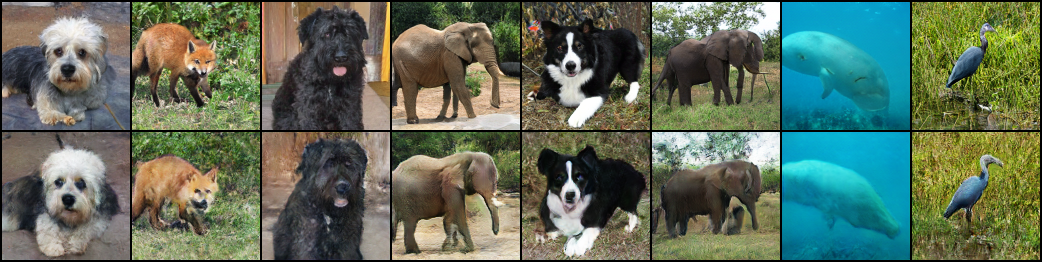}
    \end{minipage}

    \begin{minipage}{1\linewidth}
    \includegraphics[width=1\linewidth]{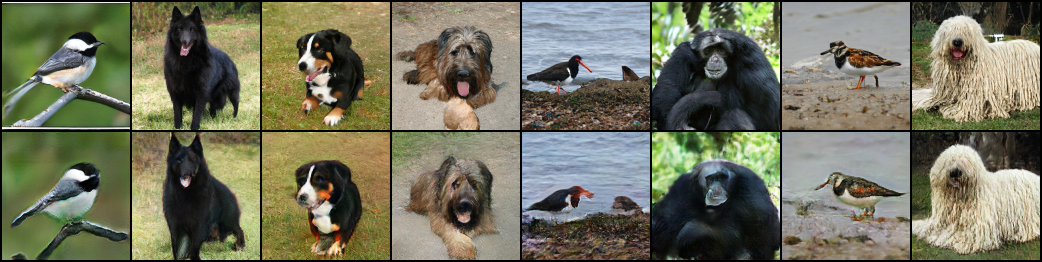}
    \end{minipage}

    \captionof{figure}{Images in the odd rows are produced by BigGAN; the even rows are generated by TinyGAN-dw given the same input.}
    \label{fig:recon_sample2}
\end{figure}

\begin{figure}[t!]
  \centering
  \includegraphics[width=1.0\linewidth]{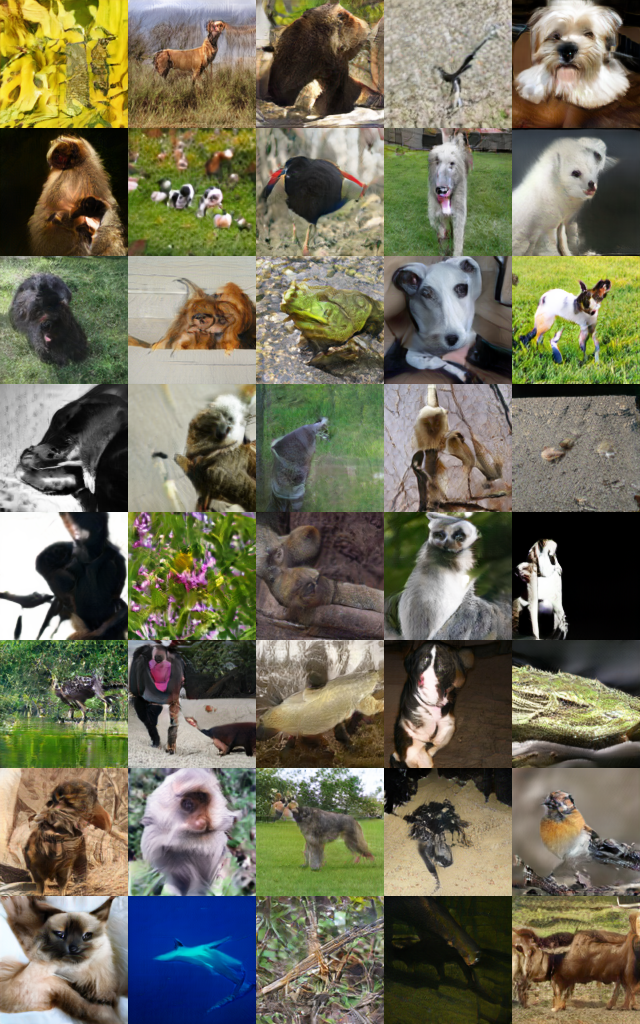}
  \caption{Randomly sampled images from SNGAN-projection.}
    \label{fig:1-sngan}
\end{figure}

\begin{figure}[t!]
  \centering
  \includegraphics[width=1.0\linewidth]{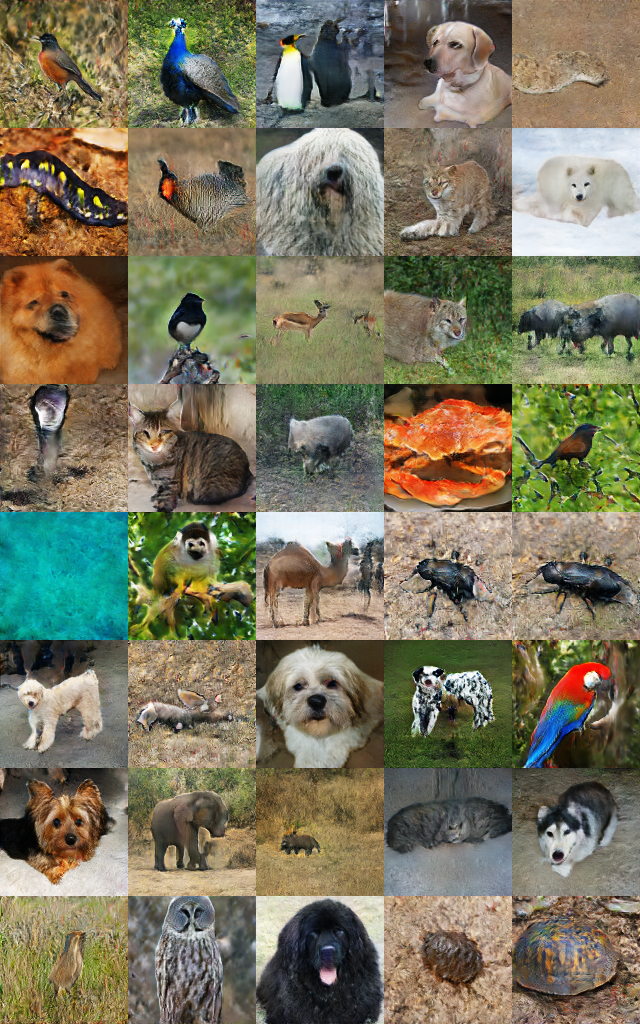}
  \caption{Randomly sampled images from TinyGAN-dw.}
\end{figure}

 \begin{figure}[t!]
  \centering
  \begin{minipage}[t]{0.48\textwidth}
  \includegraphics[width=1.0\linewidth]{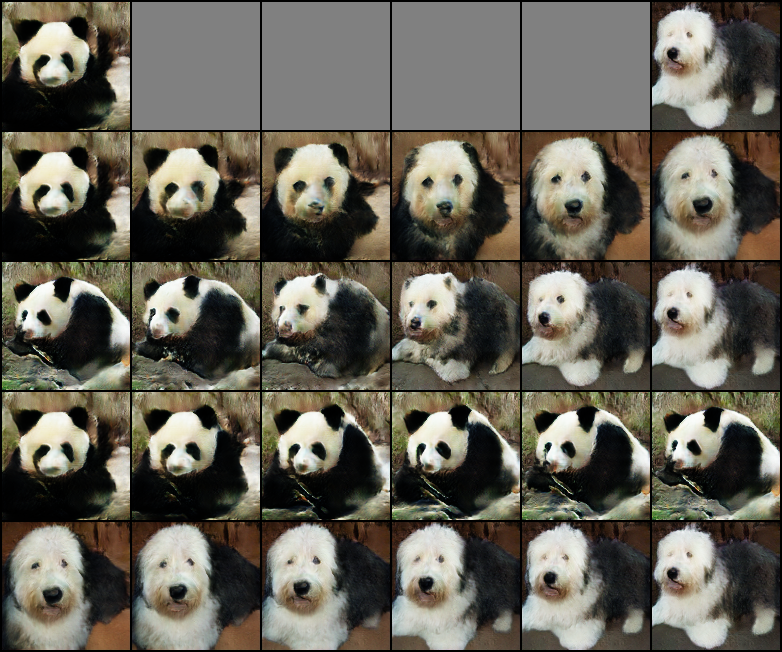}
 \end{minipage}
 \begin{minipage}[t]{0.48\textwidth}
  \includegraphics[width=1.0\linewidth]{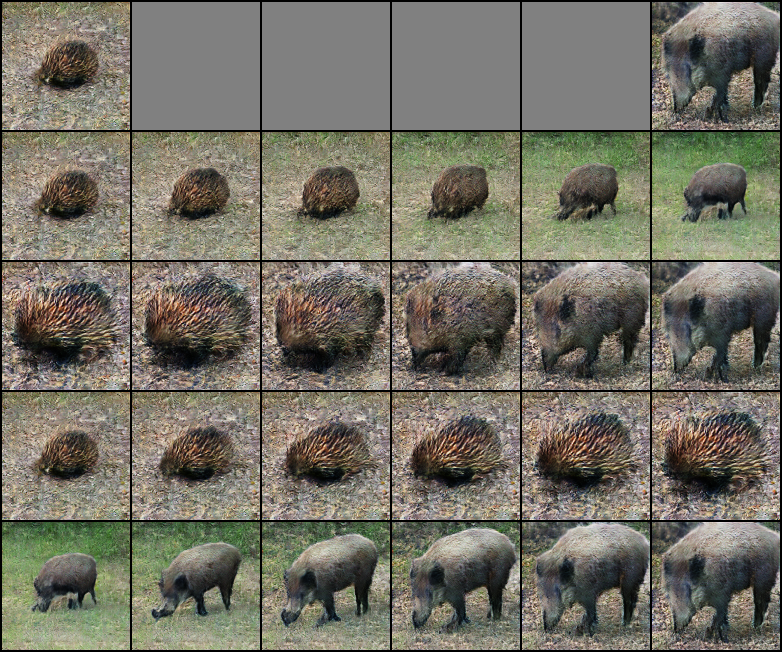}
  \end{minipage}
  
  \begin{minipage}[t]{0.48\textwidth}
  \includegraphics[width=1.0\linewidth]{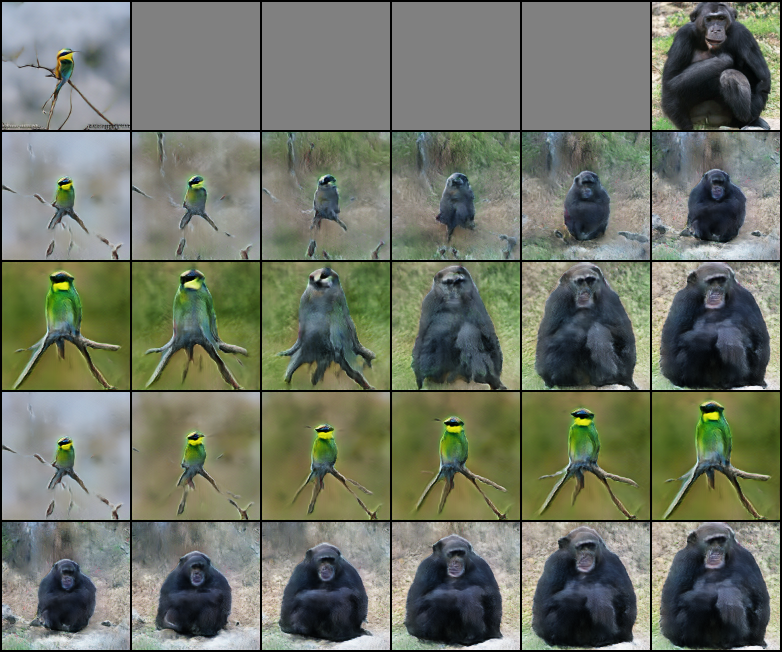}
  \end{minipage}
  \begin{minipage}[t]{0.48\textwidth}
  \includegraphics[width=1.0\linewidth]{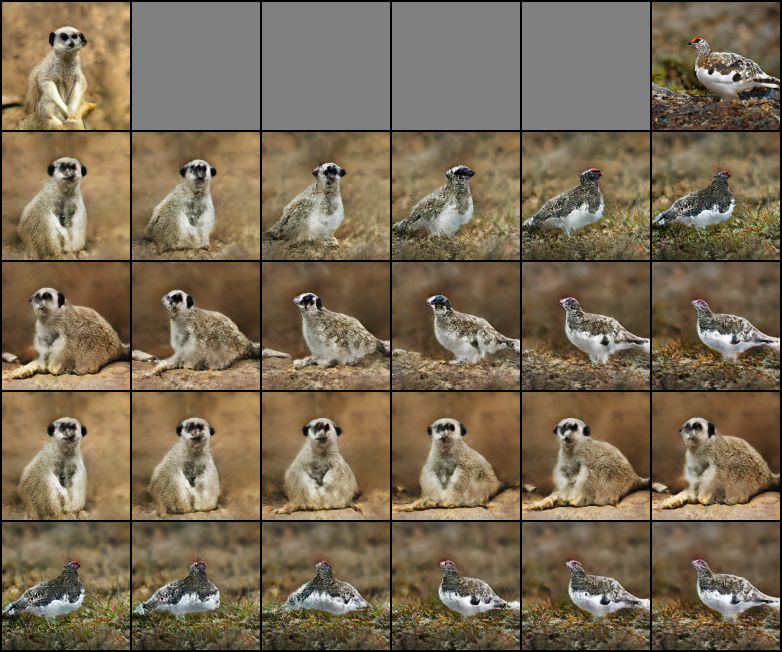}
  \end{minipage}
  \caption{Interpolations between $z, y$ pairs. The first row shows two end points generated by TinyGAN-dw. The second and third rows interpolate between $y$ with $z$ held constant, while the last two rows interpolate between $z$ with $y$ unchanged.}
\end{figure}

\clearpage
\section{Results on All Classes}
\begin{figure}[h!]
  \centering
    \begin{minipage}{1\linewidth}
    \includegraphics[width=1\linewidth]{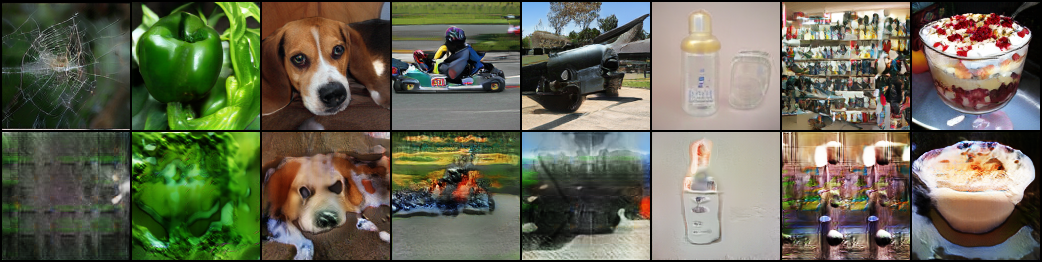}
    \end{minipage}
    
    \begin{minipage}{1\linewidth}
    \includegraphics[width=1\linewidth]{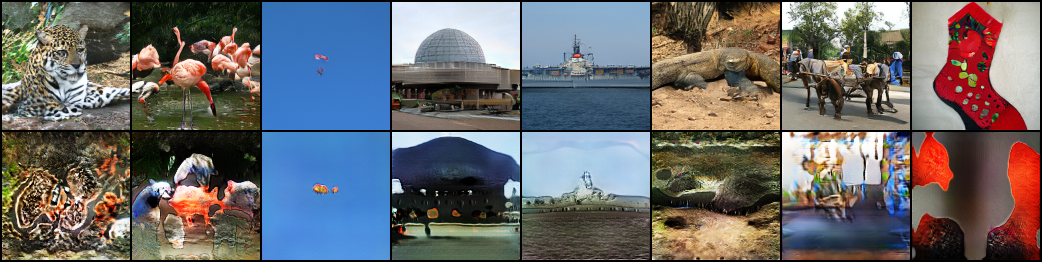}
    \end{minipage}
  \caption{TinyGAN's failure on modeling all 1000 classes in ImageNet simultaneously. Images in the odd rows are produced by BigGAN, while those in the even rows are by TinyGAN given the same input.}
\label{fig:all}
\end{figure}

While we are able to achieve satisfactory results when focusing on more homogeneous classes, such as all animal classes, we have less success when training TinyGAN on all 1000 classes of ImageNet.  
We found it hard for TinyGAN to model objects in very different categories simultaneously, especially for objects with complex textures. Figure~\ref{fig:all} shows some failure examples with blurry or unrealistic patches. 

A simple solution may be to divide the classes into a small number of groups (e.g. based on ImageNet's ontology), and train a separate TinyGAN model for each group, so that each model only needs to handle a group of more homogeneous classes. 
Another possible solution may be to use a more sophisticated training schedule and perhaps also more careful hyperparameter-tuning. 
As the focus of our work is to demonstrate the possibility of obtaining a small-size GAN with an efficient and stable training process, we leave the task of training a single small-size GAN for all the 1000 classes as our future work.

\end{document}